\documentclass[conference]{IEEEtran}
\usepackage{cite}
\usepackage{amsmath,amssymb,amsfonts}
\usepackage{algorithmic}
\usepackage{graphicx}
\usepackage{comment}
\usepackage{hyperref}
\usepackage{textcomp}
\usepackage{xcolor}
\def\BibTeX{{\rm B\kern-.05em{\sc i\kern-.025em b}\kern-.08em
    T\kern-.1667em\lower.7ex\hbox{E}\kern-.125emX}}

\usepackage{soul}   
\usepackage{framed}
\usepackage{enumitem}
\colorlet{shadecolor}{red!20} 

\IEEEaftertitletext{\vspace{-4\baselineskip}}

\begin{document}

\title{A Temporally Augmented Graph Attention Network for Affordance Classification}

\author{
\IEEEauthorblockN{
Ami Chopra, Supriya Bordoloi, Shyamanta M. Hazarika
}

\IEEEauthorblockA{
Indian Institute of Technology Guwahati, India \\
amichopra@rnd.iitg.ac.in,
bordoloi@iitg.ac.in,
s.m.hazarika@iitg.ac.in
}
}

\maketitle
\vspace{-1em}
\begin{abstract}
Graph attention networks (GATs) provide one of the best frameworks for learning node representations in relational data; but, existing variants such as Graph Attention Network  (GAT) mainly operate on static graphs and rely on implicit temporal aggregation when applied to sequential data. In this paper, we introduce Electroencephalography-temporal Graph Attention Network (EEG-tGAT), a temporally augmented formulation of GATv2 that is tailored for affordance classification from interaction sequences. The proposed model incorporates temporal attention to modulate the contribution of different time segments and temporal dropout to regularize learning across temporally correlated observations. The design reflects the assumption that temporal dimensions in affordance data are not semantically uniform and that discriminative information may be unevenly distributed across time. Experimental results on affordance datasets show that EEG-tGAT achieves improved classification performance compared to GATv2. The observed gains helps to conclude that explicitly encoding temporal importance and enforcing temporal robustness introduce inductive biases that are much better aligned with the structure of affordance-driven interaction data. These findings show us that modest architectural changes to graph attention models can help one obtain consistent benefits when temporal relationships play a nontrivial role in the task.
\end{abstract}


\section{Introduction}
Understanding and classifying neural activity related to affordance from electroencephalography (EEG) signals is a challenging problem because of intrinsic non-stationarity of neural signals, inter-subject variability, and complexity of spatiotemporal processes underlying perception, motor imagery and action execution. Early affordance research was conducted more on the neurophysiological analysis of affordances than on automated classification, as it examined the role of affordances of objects in modulating cortical activity during action observation (AO), motor imagery (MI), and their combination (AO+MI). Affordance-induced AO+MI produces greater cortical activation than AO or MI alone, and this is especially seen in ERP components such as N2 and in motor planning areas estimated using source localization techniques\cite{bordoloi2022neural}. Although these studies established the temporal and functional relevance of affordances in EEG, they relied on handcrafted analysis pipelines and did not address scalable classification.

Initial attempts of EEG classification, including the ones applied in motor imagery, and related paradigms, were dominated by handcrafted feature extraction followed by conventional machine learning classifiers. Methods like power spectral density (PSD), common spatial patterns (CSP) and band specific filtering were widely used because of their neurophysiological interpretation and neurophysiological basis. \cite{lim2025agtcnet}. However, these approaches have issues with the high dependence on expert-defined frequency bands, channel selection heuristics and task-specific assumptions, and as a result, they are difficult to generalize across subjects, sessions and experimental conditions. Moreover, spatial and temporal dimensions are treated in a rather decoupled way in such pipelines, which is inadequate for affordance-related processes that are dynamic and time-variant and involve distributed cortical interactions.

The advent of deep learning was a turning point towards end-to-end EEG representation learning. Convolutional neural networks (CNNs) gained a lot of popularity because of the capability of learning temporal and spatial filters directly from the raw signals. EEGNet is a one example, introducing compact temporal and depthwise spatial convolutions that encode EEG-specific inductive biases while maintaining a low parameter count \cite{lawhern2018eegnet}. Despite their success, CNN-based models make the implicit assumption of a grid-like structure and limits their capacity to model non-Euclidean relationships between electrodes. As a result, they have a hard time explicitly representing functional interactions between areas of the brain, which play a key role in the perception of affordances and planning of motor movements.

To solve the temporal dependencies, recurrent neural networks and temporal attention mechanism were introduced. Zhang et al. proposed a graph-embedded convolutional recurrent attention model, which assigns attention weights to the different temporal segments of EEG, which has shown improved subject-independent motor imagery \cite{zhang2020motor}. Given a sequence of temporal embeddings $\{\mathbf{z}_t\}_{t=1}^{T}$, temporal attention computes a weighted aggregation of time steps as
\begin{equation}
\beta_t = \frac{\exp(\mathbf{q}^{\top}\mathbf{z}_t)}{\sum_{k=1}^{T} \exp(\mathbf{q}^{\top}\mathbf{z}_k)},
\end{equation}
\begin{equation}
\mathbf{z} = \sum_{t=1}^{T} \beta_t \mathbf{z}_t ,
\end{equation}
where $\beta_t$ reflects the relative importance of the $t$-th temporal segment.  While such mechanisms highlight temporally salient EEG periods, spatial relationships in these models remain constrained by predefined graph structures or local convolutions, limiting representational expressiveness.
\\Graph-based representations of EEG provided a principled way to model inter-channel relationships. By treating EEG electrodes as nodes and their anatomical or functional relationships as edges, graph neural networks (GNNs) enabled explicit spatial reasoning. Surveys and empirical studies have shown that GNNs are well suited for EEG classification tasks, particularly where spatial dependencies are critical \cite{klepl2024graph} Early GNN approaches relied heavily on spectral graph convolutions, which depend on fixed graph Laplacians and do not generalize well when graph topology is uncertain or subject-specific.
\\Graph Attention Networks (GATs) addressed these limitations by replacing fixed aggregation rules with learnable attention mechanisms, allowing the model to adaptively weight neighboring nodes based on task relevance \cite{velickovic2017graph}. For a node $i$, the graph attention update at layer $l$ is given by
\begin{equation}
\mathbf{h}_i^{(l+1)} =
\sigma \left(
\sum_{j \in \mathcal{N}(i)}
\alpha_{ij}^{(l)} \mathbf{W}^{(l)} \mathbf{h}_j^{(l)}
\right),
\end{equation}
where $\alpha_{ij}^{(l)}$ denotes learned attention coefficients that enable data-driven modeling of inter-channel interactions. EEG-GAT further demonstrated that attention-based graph operators can jointly learn functional connectivity and classification-relevant representations directly from data, outperforming CNN baselines in several EEG tasks \cite{demir2022eeg}. Subsequent works, such as AGTCNet, extended this idea by jointly modeling spatial and temporal dependencies using graph-temporal attention mechanisms, emphasizing the importance of tightly coupled spatiotemporal learning \cite{lim2025agtcnet}.
\\Despite these advances, most existing GAT-based EEG models treat temporal information either through recurrent modules or fixed temporal aggregation, implicitly assuming uniform semantic relevance across time. However, affordance-related EEG signals violate this assumption, as affordances emerge, evolve, and decay over distinct temporal phases during perception, intention formation, and motor imagery. To encourage temporal robustness, temporal dropout may be introduced by randomly masking entire temporal segments during training,
\begin{equation}
\tilde{\mathbf{z}}_t =
\begin{cases}
\mathbf{0}, & \text{with probability } p, \\
\mathbf{z}_t, & \text{otherwise}.
\end{cases}
\end{equation}

In this paper, we study a temporally enhanced Graph Attention Network (GATv3) that explicitly incorporates temporal attention and temporal dropout into the GAT framework for affordance-related EEG classification. By treating time as a semantically meaningful dimension and enforcing robustness across temporally correlated observations, the proposed model aligns its inductive biases with the underlying neurocognitive structure of affordance processing. This work systematically examines how such temporal modeling improves classification performance on affordance-centric EEG data.

\section{Methodology}
\begin{figure*}[t] 
    \centering
    \includegraphics[width=\textwidth]{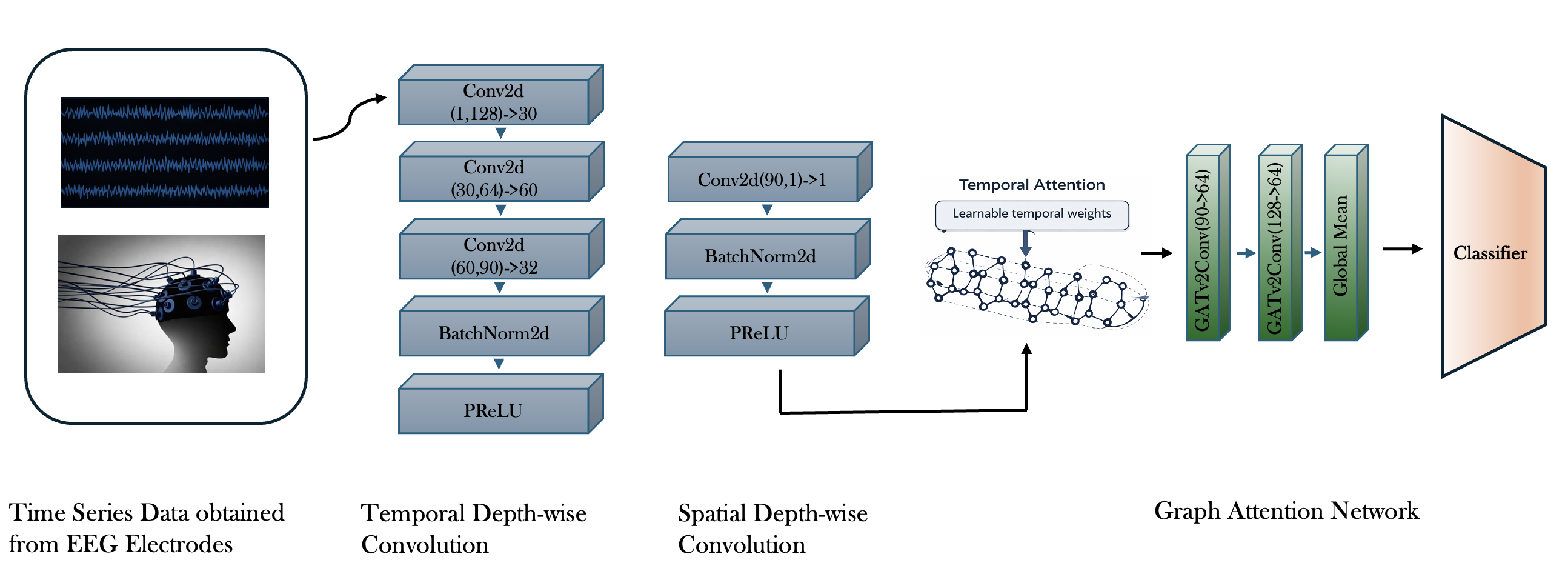} 
    \caption{EEG tGAT: Architecture of the proposed network featuring GATv2 and Temporal Attention.}
    \label{fig:network_architecture}
\end{figure*}

The primary motivation for EEG-GAT in Affordance Analysis was that Affordance-driven AO+MI tasks engage distributed cortical networks involving motor, premotor, and parietal regions, resulting in complex and dynamic inter-channel interactions \cite{bordoloi2022neural}. Conventional convolutional approaches primarily focus on temporal feature extraction and are limited in their ability to model adaptive spatial dependencies between EEG channels. By representing EEG channels as nodes in a graph and learning attention-weighted spatial interactions, the proposed EEG-GAT framework provides a principled mechanism for capturing coordinated neural dynamics underlying affordance processing.
The integration of temporal convolution, temporal self-attention, and graph attention enables the model to decode affordance-related neural activity as an emergent property of distributed and dynamically interacting cortical regions, rather than isolated EEG channels, aligning with recent advances in graph-based EEG representation learning. Fig. \ref{fig:network_architecture} shows an overview of the proposed EEG-tGAT pipeline, highlighting temporal feature extraction, temporal attention, and graph attention-based spatial modeling.

\subsection{Experimental Design and Data Acquisition}
Electroencephalography (EEG) data were acquired during an affordance-driven Action Observation and Motor Imagery (AO+MI) paradigm designed to elicit coordinated perceptual and motor cortical activity. The experimental design follows established affordance-based task paradigms in which visual stimuli depicting object-directed actions evoke neural responses associated with both action perception and internal motor simulation. Description of the dataset is explained in \cite{bordoloi2022neural}. Prior neurophysiological studies have shown that affordance-driven AO+MI engages distributed motor and premotor cortical regions more strongly than AO or MI alone, motivating its use for investigating affordance-related neural dynamics 
\subsection {EEG Data Preprocessing}
Raw EEG recordings were stored in BrainVision format and preprocessed using the MNE-Python framework. Prior to signal processing, non-essential metadata sections, including impedance information embedded in header files, were removed to ensure consistent parsing and stable data loading across recordings. Continuous EEG signals were loaded with full preloading to enable efficient in-memory processing.
To suppress power-line interference, a notch filter at 50 Hz was applied. Subsequently, band-pass filtering between 0.1 Hz and 40 Hz was performed using an infinite impulse response filter to retain task-relevant neural activity while attenuating slow drifts and high-frequency artifacts. Signals were re-referenced using a common average reference computed across all EEG channels, excluding the vertical electrooculogram (VEOG) channel, following standard EEG preprocessing practices adopted in prior EEG classification studies \cite{lawhern2018eegnet}, \cite{zhang2020motor}.
Independent Component Analysis (ICA) was incorporated as an optional artifact correction step. When enabled and when VEOG channels were available, ICA components correlated with ocular activity were automatically identified using EOG-related epochs and removed. This conditional application ensured effective eye-movement artifact suppression without enforcing uniform component rejection across all recordings, aligning with practical EEG preprocessing strategies in subject-variable datasets \cite{zhang2020motor}.
\\All EEG signals were resampled at 256 Hz in order to ensure uniform temporal resolution across subjects and trials. Task-related EEG epochs were extracted from the data using experimental annotations of the events. Only task-relevant markers that corresponded to motor imagery and control conditions were retained and rest and unrelated annotations were excluded. For each valid event, EEG data were obtained from 9s to 15s after the onset of the stimulus, so that a six-second window centered around sustained engagement in a task was obtained.
Each epoch was normalized on a per channel basis by z score normalization to minimize the inter trial amplitude variability in order to get more numerical stability during model training. The normalized epochs were further divided into non-overlapping one-second temporal windows resulting in several temporally-localized samples from each trial.

\subsection{Graph-based EEG Representation}
Each segment of one second of EEG was represented as a graph in order to explicitly model spatial dependencies between EEG channels. In this formulation, each of the EEG channels was regarded as a node, and node features comprised the temporal EEG signal in the corresponding channel segment. A fully connected undirected graph was developed for each EEG segment, where all the channels could interact depending on the lack of predefined anatomical and functional connectivity constraints.
This choice of design is inspired by the recent graph-based EEG classification methods where hand-engineered adjacency matrices are avoided and the model is given the opportunity to learn task-wise inter-channel relationships directly from data \cite{demir2022eeg}.  Such flexibility is especially important in affordance driven paradigms, where functional connectivity patterns may change between experimental conditions and subjects.
Each of these EEG one-second segments was considered as independent samples of graph in training. This approach enlarges the effective number of training samples, and allows the model to learn temporal and spatial representations together in the unified graph learning framework \cite{klepl2024graph}
\subsection{Temporal Feature Extraction Network}
To capture temporal EEG dynamics on each channel, the node features were first processed with a temporal convolutional architecture based on compact EEG-specific CNN architectures such as EEGNet \cite{lawhern2018eegnet}. Three successive two-dimensional convolutional layers were applied along the temporal dimension with kernel sizes of 1 x 128, 1 x 64 and 1 x 32, which corresponds to the different temporal resolutions of the sampling frequency of 256 Hz. This multiscale temporal processing allows the network to include low-frequency as well as high-frequency neural oscillations that are relevant for motor imagery and affordance perception.
Each convolutional layer was followed by batch normalization and parametric rectified linear unit (PReLU) activation. Spatial dropout was used to decrease the overfitting and to help the model generalize from subject to subject. Following temporal convolution, a depthwise convolution was applied over the channel dimension as a spatial filter, which preserves channel specific temporal features, while also allowing for localized spatial integration before graph-based modeling, in line with previous EEGNet-inspired architectures \cite{lawhern2018eegnet}.
To further improve temporal modeling, a temporal self attention mechanism was used over the convolutional feature maps. This module is responsible for computing attention weights along the temporal dimension, and can be used to make the network adaptively focus on task-relevant time points within each EEG segment. Temporal features were then aggregated with mean pooling over time, which resulted in a fixed-length feature vector for every EEG channel. The inclusion of a temporal attention is motivated by previous studies showing effectiveness of this in highlighting discrimination of temporal periods in EEG signals \cite{zhang2020motor}.
\subsection{Graph Attention-Based Spatial Modeling}
Spatial dependencies between EEG channels were modeled by Graph Attention Networks based on the GATv2 convolutional operator. Channel wise feature vectors derived from temporal processing were taken as node embeddings and fed into two successive GATv2 layers. The first layer used multi-head attention with feature concatenation, which enables the model to focus on multiple interaction subspaces at once. The second layer consisted of a single attention head to generate small and stable node representations for graph-level aggregation.
Graph attention mechanisms allow for the adaptive weighting of the inter-channel interactions based on data and has demonstrated an improvement over fixed or heuristic graph constructions for EEG classification tasks \cite{vrahatis2024graph}, \cite{demir2022eeg}. Layer normalization was used after each GATv2 layer in place of the batch normalization, in order to obtain better stability under different batch sizes and to reduce subject-specific distributional shifts in graph-based EEG learning. PReLU activation was applied after each normalized graph attention layer which brings nonlinearity while keeping the gradient flow stable.
\subsection{Graph Level Readout and Classification}
In order to get a segment-level representation, node embeddings generated by the last graph attention layer were combined using global mean pooling across all EEG channels. This readout operation produces a graph-level embedding that summarizes the distributed cortical activity for each segment of the EEG, which is in line with common graph classification pipelines \cite{demir2022eeg}.
The resultant embedding was fed into a multilayer perceptron classifier which has a fully connected layer with ELU activation formula, dropout regularization and a final linear layer to produce class logits corresponding to the experimental conditions.
\subsection{Model Training and Evaluation}
The EEG-GAT model has been trained with the AdamW optimizer with an initial learning rate of $3 \times 10^{-4}$ and a weight decay of $1  \times  10^{-3}$. Cross-entropy loss with label smoothing was used as an objective function in order to mitigate overconfidence and better generalization. Training was done by mini batch stochastic optimization early stopping based on validation loss to avoid over fitting. A Reduce-on-Plateau learning rate scheduler was applied to adaptively reduce the learning rate when there was no improvement in validation performance.
Model performance was tested with a 5-fold cross validation where samples of EEG graphs were grouped using trial ID's in each of the folds in order to avoid data leakage in test and train sets. Each segment of EEG data of one second was considered as an independent graph for training and evaluation, which allowed the model to learn generalized spatio-temporal representations across trials and subjects.

\begin{figure*}[t]
    \centering
    \includegraphics[width=0.3\textwidth]{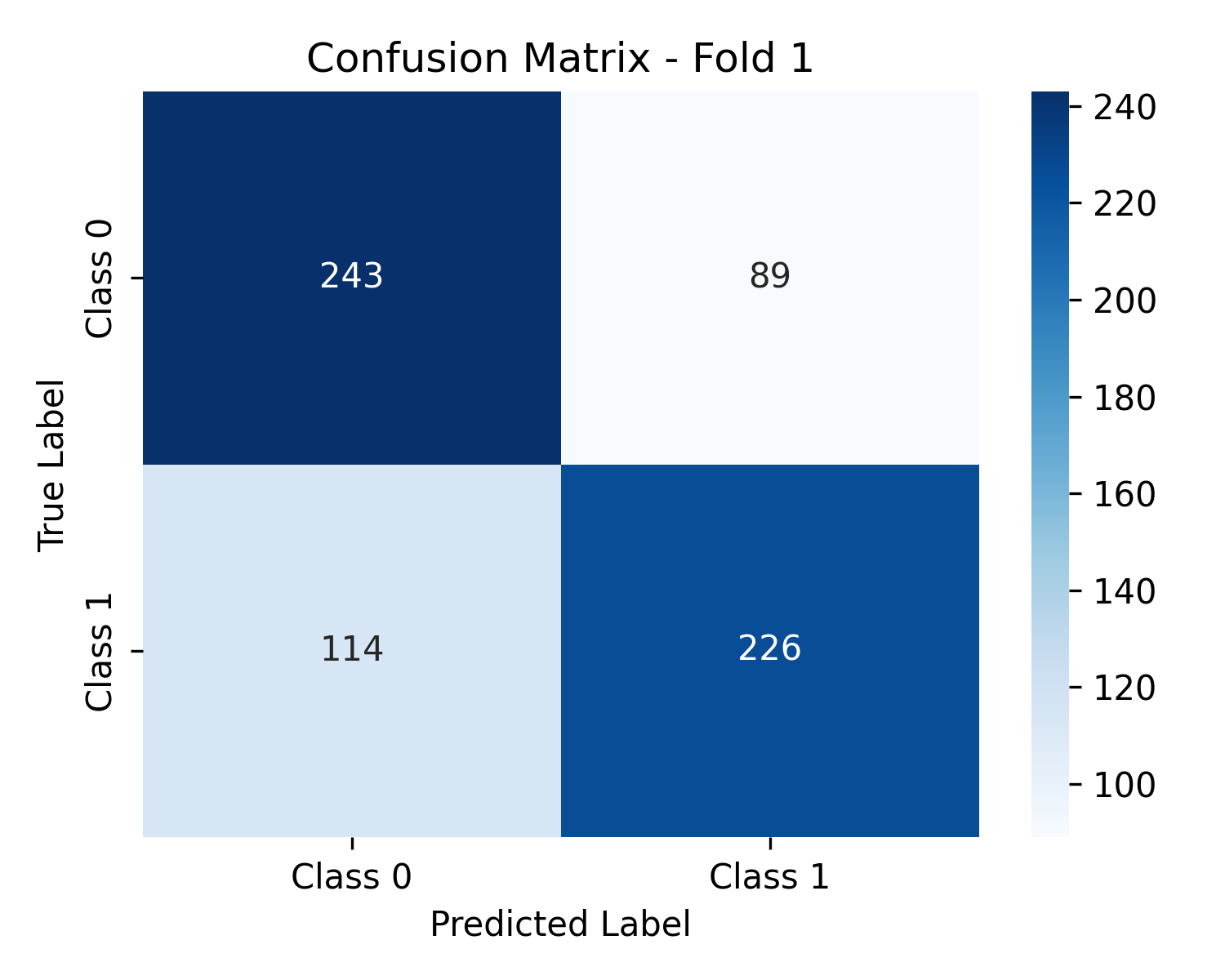}
    \hspace{1cm} 
    \includegraphics[width=0.3\textwidth]{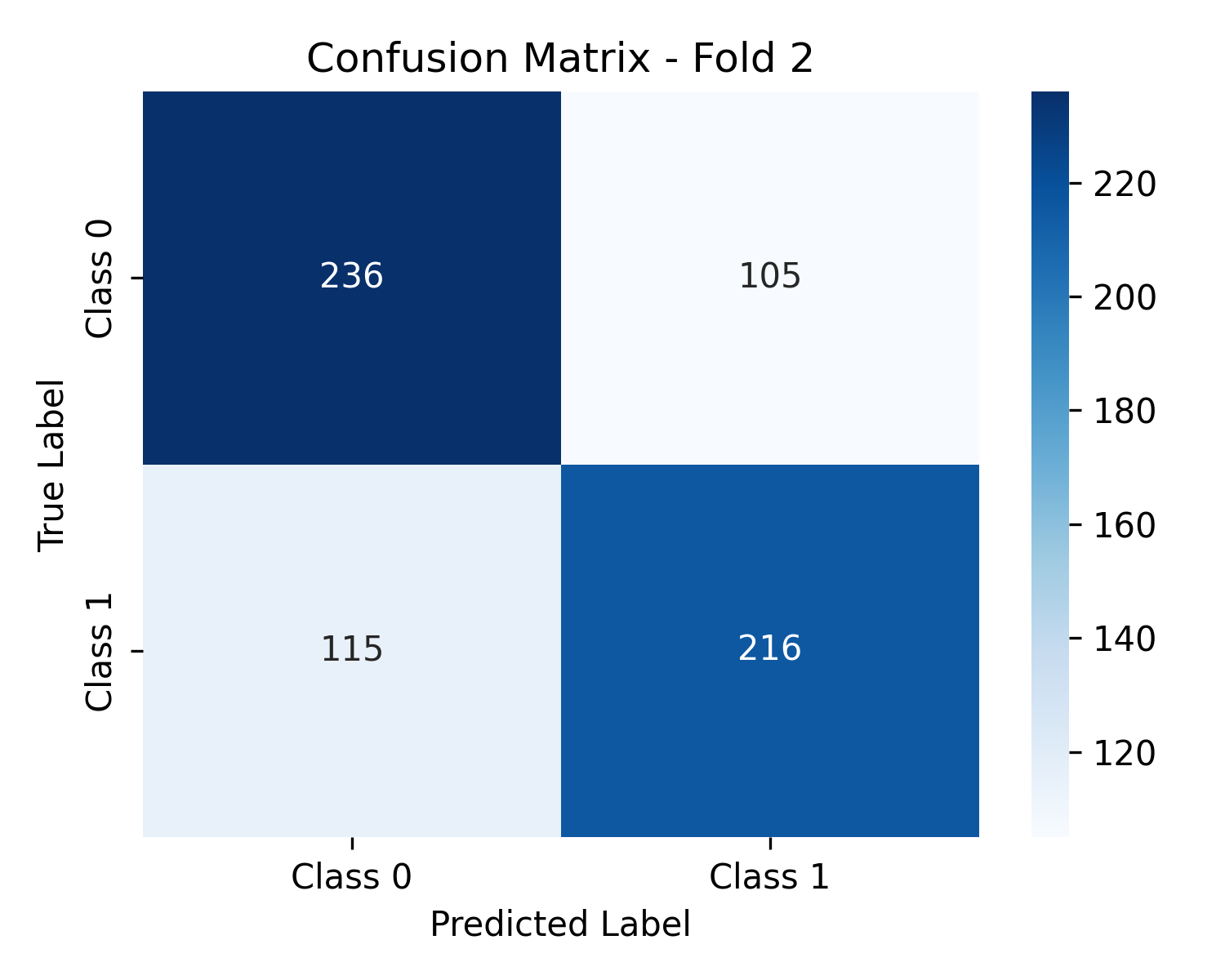}
    
    \vspace{0.4cm} 

    \includegraphics[width=0.30\textwidth]{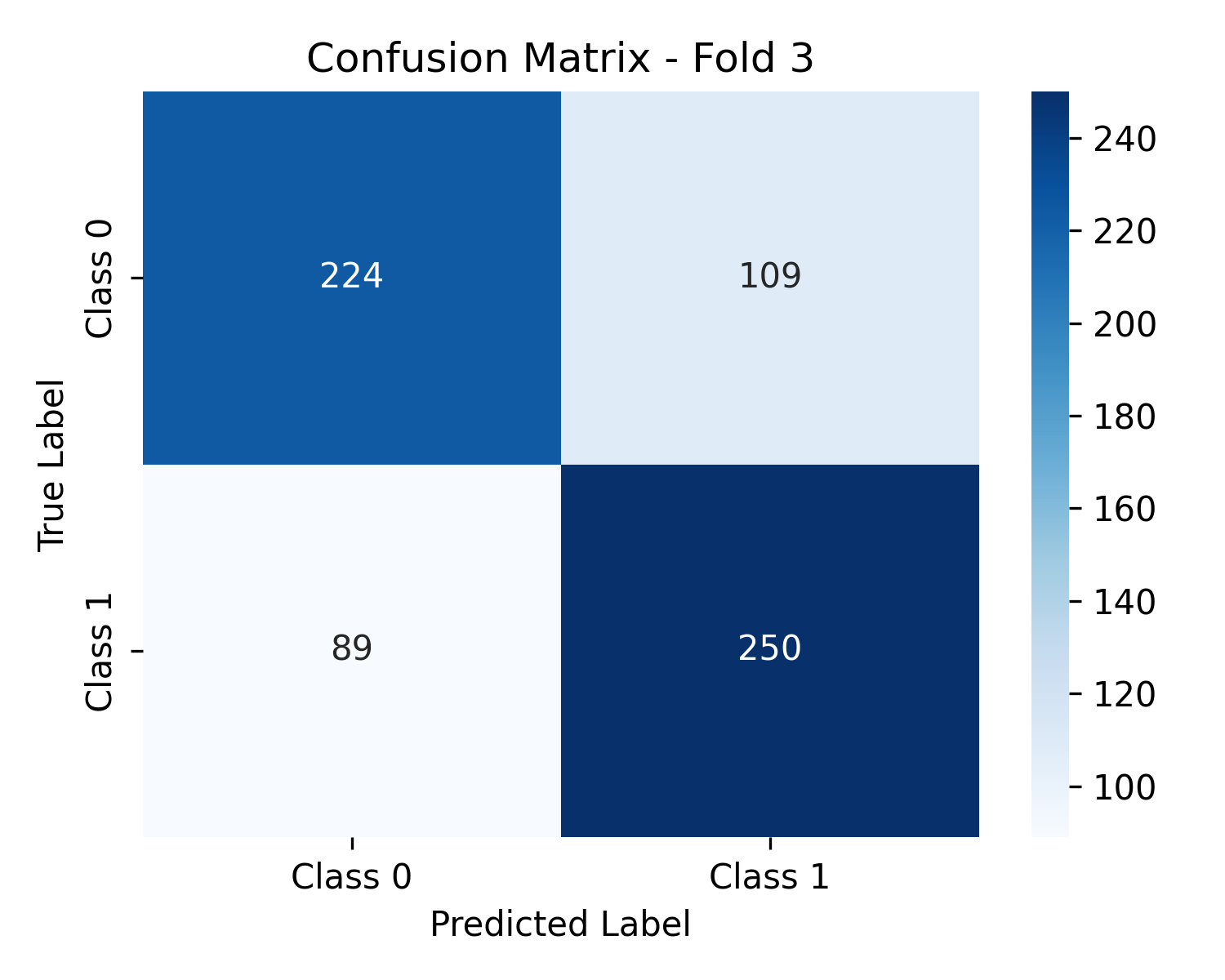}
    \hfill
    \includegraphics[width=0.30\textwidth]{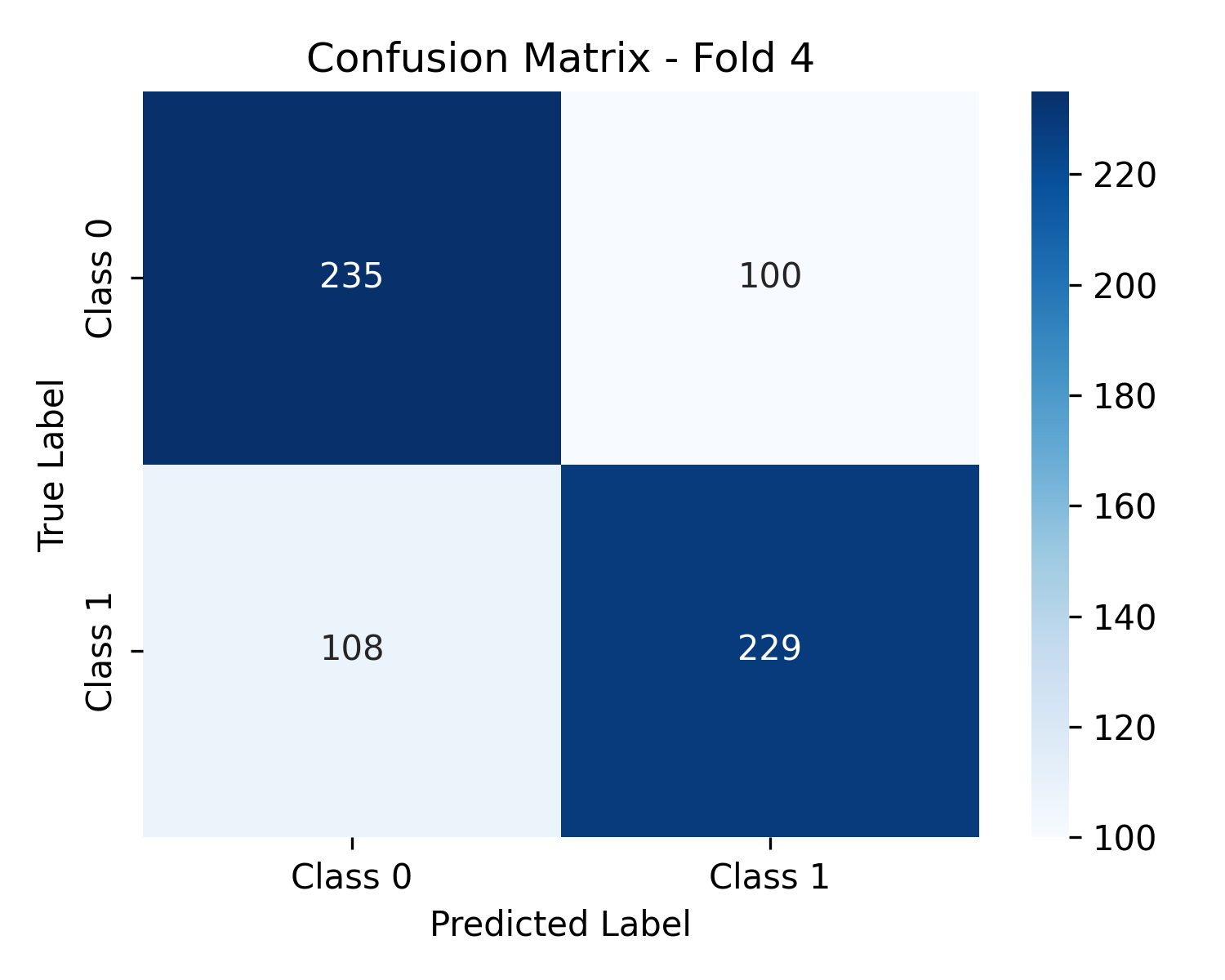}
    \hfill
    \includegraphics[width=0.30\textwidth]{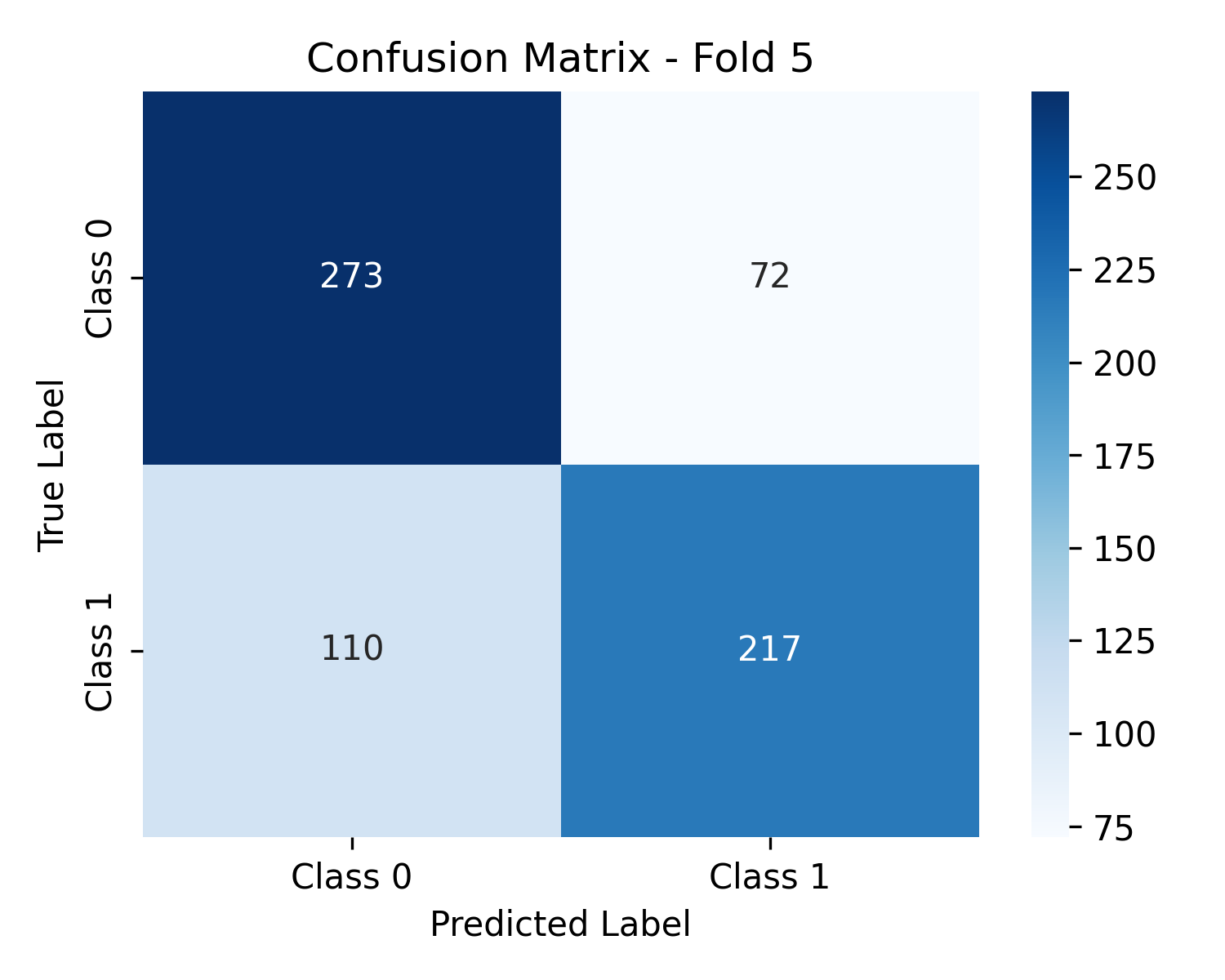}
    
    \caption{Confusion matrices for all five cross-validation folds.}
    \label{fig:confusion_matrices}
\end{figure*}

\section{Results}

The proposed model has been tested with 5-fold cross validation to test the robustness and generalization capability of the model. In each fold, the dataset was split into an exclusive training and test set, and the model selection process was done using early stopping based on validation loss. Performance is reported using accuracy, precision, recall, F1-score and Cohen's Kappa.

A full ablation study has not been performed in this work because of the closely integrated design of temporal attention and temporal dropout as well as the associated retraining cost since the goal of the evaluation is to measure overall effectiveness and stability of the proposed temporally informed graph attention framework instead of to isolate the contribution of single architectural parts. Instead, results are presented in the form of complementary quantitative results and visual analyses that together convey knowledge about classification performance and class-wise behavior, as well as cross-fold consistency.

\subsection{Performance Analysis}

On the 5 folds, the model obtained an average classification accuracy of $71.55\% \pm 1.58\%$ with closely matched precision ($71.61\% \pm 1.60\%$), recall ($71.52\% \pm 1.55\%$), and F1-score ($71.51\% \pm 1.56\%$). The low variance across folds suggests stable behavior of learning under different partitions of data. Cohen's kappa averaged $0.43$ indicating moderate levels of agreement above and beyond chance and confirming that the observed performance is not due to class imbalance.

These results show that the model has a balanced class-wise performance, as shown by the close correspondence between the precision and recall. The consistency of the metrics between folds is also a further indication of reliable generalization of the learned representations across different subsets of the data.

\subsection{Comparison with Baseline Models}
To evaluate the effectiveness of EEG-tGAT we compared it against several commonly used baseline models, including EEGNet, CNN-LSTM, and a spatial GATv2 model without temporal augmentation. 
The EEGNet model achieved an average accuracy of 54.88\% $\pm$ 1.04\%, indicating limited capability in capturing complex temporal dependencies present in affordance-driven EEG signals. The CNN-LSTM architecture achieved improved performance with an average accuracy of 70.36\% $\pm$ 5.46\%, reflecting the advantage of incorporating sequential temporal modeling. The GATv2 model achieved 69.91\% $\pm$ 1.40\%, demonstrating the benefit of graph-based spatial learning compared to convolution-only methods.
In comparison, the proposed EEG-tGAT model achieved the highest performance with 71.55\% $\pm$ 1.58\% outperforming all baseline configurations.
\subsection{Ablation Study}
An ablation study was conducted by selectively removing temporal dropout, temporal attention, and both components together. The full EEG-tGAT model served as the reference configuration with an accuracy of 71.55\% $\pm$ 1.58\%.
When temporal dropout was removed, classification performance decreased to 68.60\% $\pm$ 1.84\%, showing that dropout contributes significantly to improving robustness against temporally correlated noise and reducing overfitting during training. Removing temporal attention resulted in an accuracy of 70.53\% $\pm$ 2.58\%, showing that temporal weighting improves the model’s ability to focus on informative temporal segments within EEG signals. When both temporal attention and temporal dropout were removed simultaneously, the model achieved 69.91\% $\pm$ 2.49\% confirming that the combined presence of these components provides complementary benefits.
\\Overall, the consistent performance degradation observed across all ablation configurations show us that both temporal attention and temporal dropout contribute meaningfully to the stability and discriminative capability of the proposed EEG-tGAT architecture.

\subsection{Confusion Matrix Analysis}

To study the class-wise prediction behavior in more detail, confusion matrices were calculated for each fold using the best performing model checkpoint. Confusion matrices for all five folds are shown in Fig.~\ref{fig:confusion_matrices}.

An inspection of the per fold confusion matrices shows a consistent structure of errors between folds. In the case of both classes, true positiveness and true negativeness lead the diagonal entries, while misclassifications are fairly distributed between false positive and negative. This balance translates to the good similarity between precision and recall values, which shows that the model is not particularly biased towards one class or the other.

Higher overall accuracy folds (e.g., Fold~5) also exhibit less off-diagonal confusion which implies better temporal and relational discrimination than class specific overfitting. Despite these improvements, the confusion matrices show a persistent amount of overlap between the two classes with misclassification rates being non-negligible between all of the folds. This is a behavior that is consistent with the inherent difficulty of the task, and the noisy and temporally distributed nature of EEG-based affordance signals.

\subsection{Stability Analysis of Cross Fold Stability}

\begin{figure}[t]
    \centering
    \includegraphics[width=0.7\columnwidth]{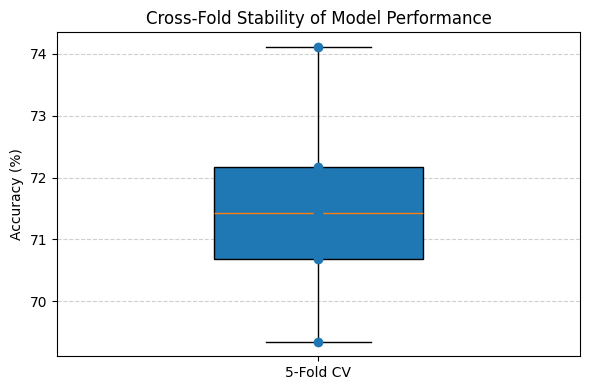}
    \caption{Box plot showing distribution of classification accuracy across five folds.}
    \label{fig:boxplot_accuracy}
\end{figure}

To determine the robustness, the distribution of classification accuracy for folds is plotted in the form of box plot as shown in Figure~\ref{fig:boxplot_accuracy}. The small interquartile range and lack of extreme outliers suggest that there are limited variations in the folds of performance.

Overall, these results show that our approach provides consistent and reproducible gains across folds, indicating the effectiveness of incorporating temporally informed graph attention mechanisms for affordance classification. 

\section{DISCUSSION}
The performance improvements achieved with the proposed EEG-tGAT architecture, which is based on the temporally augmented EEG-GAT architecture can be attributed to its explicit handling of the spatiotemporal dynamics that are inherent in affordance-based EEG signals. Standard GAT architectures are mostly focused on modeling relational structure between nodes at a single time instant. Although attention weights are learned in an adaptive manner, the formulation implicitly assumes that node features at a given snapshot are sufficient in order to discriminate. By introducing the temporal attention, the proposed model is able to explicitly capture this temporal evolution instead of compressing this into static node embeddings. This is especially true for motor imagery and action observation paradigms, in which affordance-related neural patterns develop, peak, and then decay over different intervals of time. \\Some parts, such as the start of motor priming or coherence over extended periods of imagery, will have a stronger semantic relevance as compared to others which may be dominated by background activity or noise. Temporal attention helps the model to selectively focus on such informative periods and de-emphasize the effect of less informative period of time by reducing the averaging effects of models trained by treating time uniformly. Temporal dropout is then added to further complement this mechanism by adding regularization along the temporal dimension. EEG signals have strong temporal autocorrelation, and without temporal regularization the models might overfit to any coincidences that last only a short time, without being robust to trials/subjects. Not only does it encourage the network to rely on the information dispersed throughout the entire epoch, but not to memorize narrow windows of time. This property is of particular importance in affordance-driven EEG analysis in which meaningful neural representations must therefore be stable even in the presence of noisy and partially corrupted individual segments.\\Importantly, temporal attention and temporal dropout interact in a mutually reinforcing manner. While the attention mechanisms alone can render themselves too selective and focus too much on a few frames, dropout helps to prevent premature collapse of attention weights.\\The temporal weighting result is therefore smoother and more generalized, capturing both salient instants, as well as longer range temporal dependencies. From a representation learning point of view, the proposed approach is able to separate structural importance (corresponding to inter-channel relationships) and temporal importance (corresponding to the timing and persistence of neural activation). This separation creates an inductive bias that is more consistent with the physiological characteristics of affordance processing. Considering time as a meaningful dimension instead of a noisy variable, the learning process of the model is directed toward learning stable neural correlates of affordance-related cognition, rather than superficial correlations. This correspondence between model design and data structure gives a principled explanation of the consistent performance improvements in the experimental evaluation.

\section{CONCLUSION}
In this work, we present EEG-tGAT, an enhanced graph attention network for EEG classification that integrates explicit temporal attention and dropout regularization. By jointly modeling spatial relationships among EEG channels through learned graph attention and emphasizing informative temporal segments, the proposed approach effectively captures spatio-temporal dependencies inherent in EEG signals.\\Experimental evaluation using five-fold cross-validation demonstrated that EEG-tGAT model achieved a mean classification accuracy of $71.55\% \pm 1.58\%$, with closely matched precision ($71.61\% \pm 1.60\%$), recall ($71.52\% \pm 1.55\%$), and F1-score ($71.51\% \pm 1.56\%$) The use of early stopping further ensured stable generalization and mitigated overfitting, despite increasing training accuracy in later epochs. These results indicate that incorporating temporal attention within a graph-based framework yields more robust and discriminative representations than spatial-only GAT variants.
\\Overall, EEG-tGAT offers a balanced trade-off between model complexity and performance, achieving competitive results while maintaining training stability. Future work will focus on subject-independent evaluation, deeper temporal modeling, and extending the framework to multi-task and multi-modal EEG analysis scenarios.

\bibliographystyle{IEEEtran}
\bibliography{ref}

\end{document}